\def\year{2018}\relax
\documentclass[letterpaper]{article} 
\usepackage{aaai18}  
\usepackage{times}  
\usepackage{helvet}  
\usepackage{courier}  
\usepackage{url}  
\usepackage{graphicx}  
\usepackage{amsmath}
\usepackage{bm}
\usepackage{multirow}
\begin{document}
%
\title{Multi-Channel Pyramid Person Matching Network for Person Re-Identification}

\author{$\text{Chaojie Mao}^{1}$, $\text{Yingming Li}^{\thanks{Corresponding author},1}$, $\text{Yaqing Zhang}^{1}$, $\text{Zhongfei Zhang}^{1}$, and $\text{Xi Li}^{2, 3}$\\
${}^{1}$ College of Information Science \& Electronic Engineering, Zhejiang University, Hangzhou, China\\
${}^{2}$ College of Computer Science and Technology, Zhejiang University, Hangzhou, China\\
${}^{3}$ Alibaba-Zhejiang University Joint Institute of Frontier Technologies, Hangzhou, China\\
\{mcj, yingming, zhongfei, yaqing, xilizju\}@zju.edu.cn \\
}

\maketitle

\begin{abstract}
In this work, we present a Multi-Channel deep convolutional Pyramid Person Matching Network (MC-PPMN) based on the combination of the semantic-components and the color-texture distributions to address the problem of person re-identification. In particular, we learn separate deep representations for semantic-components and color-texture distributions from two person images and then employ pyramid person matching network (PPMN) to obtain correspondence representations. These correspondence representations are fused to perform the re-identification task. Further, the proposed framework is optimized via a unified end-to-end deep learning scheme. Extensive experiments on several benchmark datasets demonstrate the effectiveness of our approach against the state-of-the-art literature, especially on the rank-1 recognition rate.
\end{abstract}

\section{Introduction}
The task of person Re-Identification (Re-ID) is to judge whether two person images indicate the same target or not and has widespread applications in video surveillance for public security. From the perspective of human perception, two persons can be distinguished according to the color or texture features of the persons' attributes (e.g. clothes, hairs) and the latent semantic parts (e.g., head, front and back upper body, belongings). Consequently, the person Re-ID task can be addressed from two aspects’ matching: the color-texture distributions and the latent semantic-components.

\begin{figure}[htb!]
\centering
\includegraphics[width=0.43\textwidth]{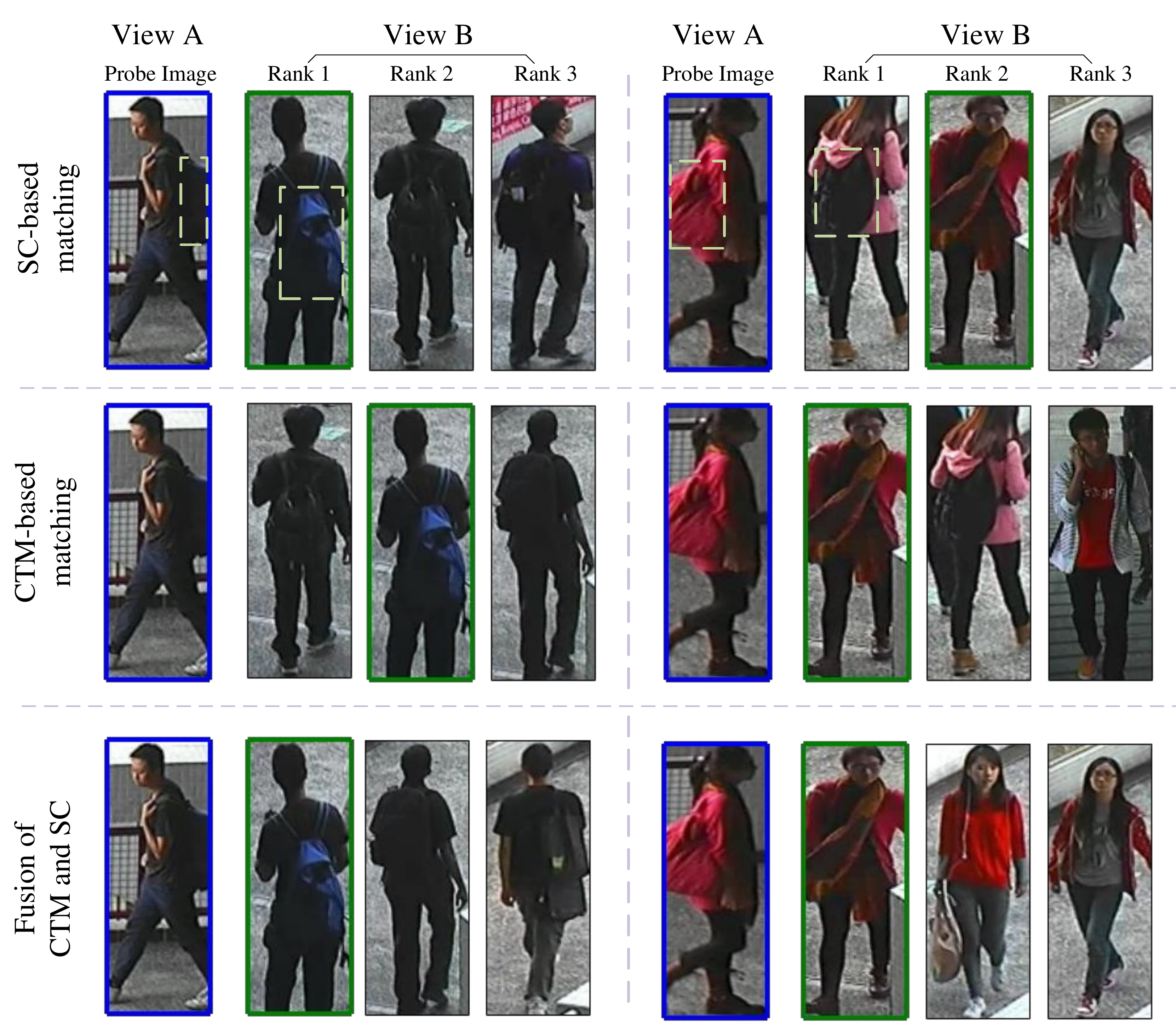}
\caption{Example pairs of images from the CUHK03 dataset. Given the probe image
of a person in view A marked by a blue window, the task is to find the same person in the gallery set of view B. The groundtruth images are marked by the green bounding boxes. The first row and the second row are re-identification results by semantic components (SC)-based and color-texture maps (CTM)-based strategies, respectively. Failures exist in both cases. The third row are the results by the combination of the two strategies which obtain success on both examples. }
\label{fig:realexamples}
\end{figure}

In the previous efforts, two common strategies are employed for the person Re-ID task. One strategy focuses on learning the correspondence among color-texture distributions from different person images, but ignoring correspondence among the semantic-components~\cite{Alpher11}~\cite{Alpher12}~\cite{Alpher13}~\cite{Alpher14}. The other relies on learning the correspondence among semantic-components, while ignoring the color-texture correspondence~\cite{Alpher16}~\cite{Alpher27}~\cite{Alpher19}~\cite{Alpher06}~\cite{Alpher20}~\cite{Alpher21}~\cite{Alpher22}. Figure \ref{fig:realexamples} gives two examples to show respective advantages of the two strategies, where images in the first row are re-identified results by the semantic correspondence and images in the second row are re-identified results by the color-texture correspondence..

In this work, we assume that the semantic-components and color-texture distributions are complementary to each other and present a novel multi-channel deep convolutional person matching network based on the combination of the semantic-components and the color-texture distributions. In particular, we learn separate deep representations for semantic-components and color-texture distributions from two person images and then employ the matching network to obtain the correspondence representations. These correspondence representations are fused to address the Re-ID task.

On one hand, to learn the correspondence among semantic-components from two persons, we first fine-tune the model weights of the ImageNet-pretrained GoogLeNet~\cite{Alpher09} to learn the deep representation of each person's semantic-components. By visualizing some layers of this network, we observe that the discriminative regions in feature maps correspond to different components (bag, head, body, etc.) of a person. For matching these learned feature regions from two person images, convolution operation is exploited to fuse these feature regions from different inputs in the same sub-windows. However, the feature regions of the same components from two views for one person seldom have the consistent spatial scale and location due to viewpoint changes. To overcome the variation of spatial scale and location, we employ atrous convolution ~\cite{Alpher23} with multi-scale views to construct a module called pyramid matching module, which provides a desirable view of perception without increasing parameters and computation by introducing zeroes between the consecutive filter values. With this module, we obtain the correspondence representation between the semantic-components from different inputs.

On the other hand, to build the correspondence representation between color-texture distributions, we propose to introduce the deep color-texture distribution representation learning based on convolutional neural network. Different from the conventional hand-crafted features (e.g., LOMO)~\cite{Alpher32}, we first extract RGB, HSV and SILTP histograms ~\cite{Alpher37} with the sliding windows and then project the histogram bins into specific feature maps, which encode the spatial distribution for the particular color-texture range. With these Color-Texture feature Maps (CTM), we employ a 3-layers convNet to learn the deep color-texture representation for each person image. Thus, the pyramid matching module is exploited to learn the correspondence representation between color-texture distributions from different person images.

Having the learned correspondence representations for the semantic-components and color-texture distributions, the MC-PPMN is carried out by fusing them with two fully connected layers to decide whether the two input images represent the same person or not. The proposed framework is evaluated on several real-world datasets. Extensive experiments on these benchmark datasets demonstrate the effectiveness of our approach against the state-of-the-art, especially on the rank-1 recognition rate.

The main contributions of this paper are as follows:

(1) We propose a deep convolutional network named MC-PPMN which learns the correspondence representations from both the semantic-components and color-texture distributions. Deep structures for encoding both semantic space and color-texture distributions, and cross-person correspondence are jointly optimized to improve the generalization performance of the person re-identification task.

(2) The proposed framework employs the pyramid matching strategy based on the atrous convolution to learn the correspondence representation for two person images, which provides a desirable view of perception without increasing parameters and computation by introducing zeroes between the consecutive filter values.

\begin{figure*}[htb!]
\includegraphics*[width=\textwidth]{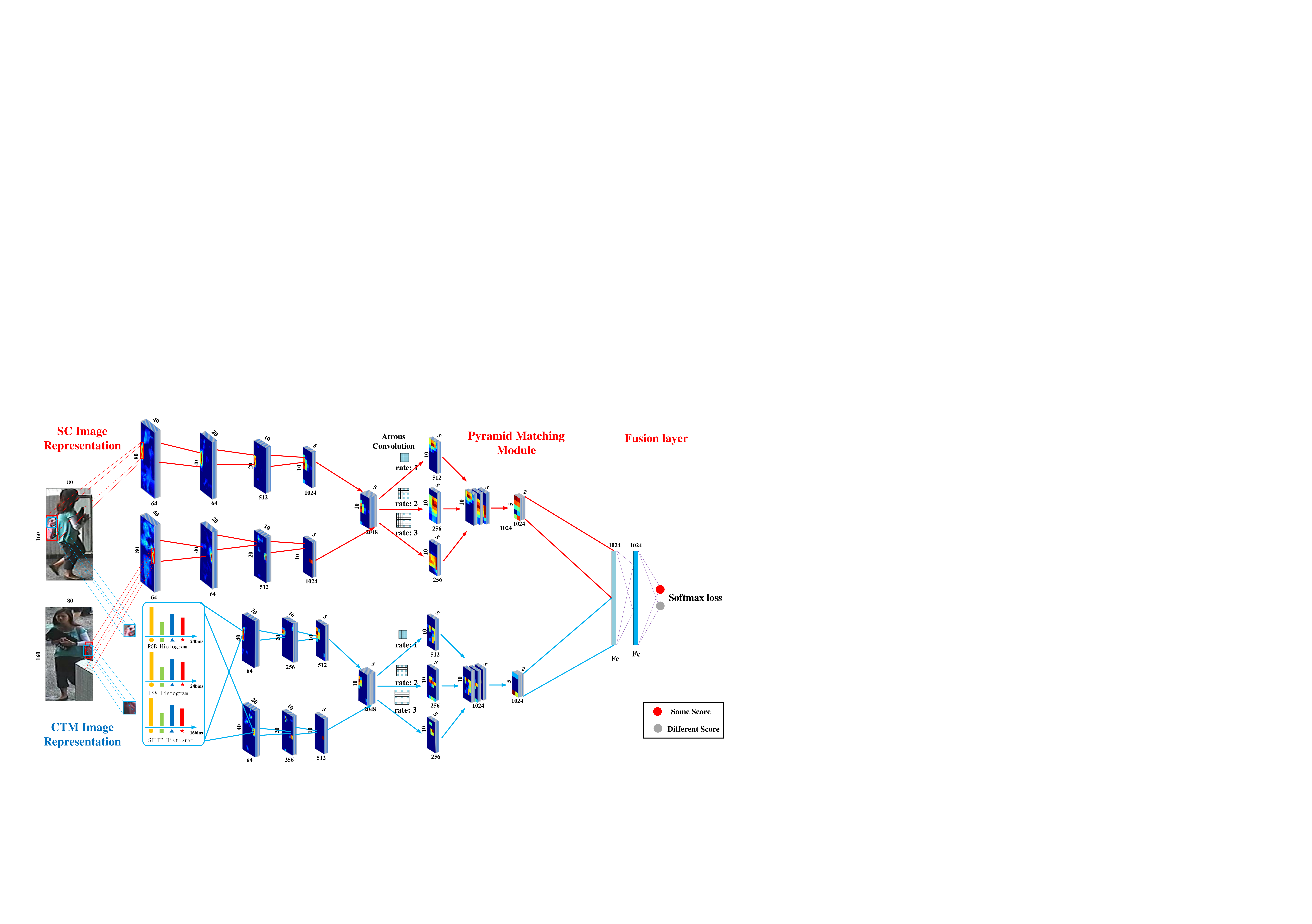}
\caption{The proposed architecture of deep convolutional person matching network. }
\label{fig:architecture}
\end{figure*}

\section{Related Work}
In the past five years,  many efforts have been proposed for the task of person Re-ID, which greately advance this field. The discrimative feature representation learning and the effective matching strategy learning are the main topics for person Re-ID. For feature representation, many approaches design the robust descriptors againist misalignments and variations with color and texture, which are two of the most useful characteristics in image representation.  The hand-crafted features including HSV color histogram~\cite{Alpher01}, SIFT histogram~\cite{Alpher02}, LBP histogram~\cite{Alpher03} features and the combination of them are widely used for image representation. Many efforts also consider the properties of person images such as  symmetry structure of segments~\cite{Alpher01} and the horizontal occurrence of local features\cite{Alpher32}, to design the features, which significantly boost the matching rate. 

For the matching strategy, the metric learning is the basic idea to find a mapping function from the feature space to the distance space so as to minimize the intra-personal variance while maximizing the inter-personal margin. Many approaches have been proposed based on this idea including pair-wise constrained component analysis (PCCA)~\cite{Alpher11}, local Fisher discriminant analysis (LFDA)~\cite{Alpher12}, Large Margin Nearest-Neighbour (LMNN)~\cite{Alpher13}, and KISS metric learning (KISSME)~\cite{Alpher14}. However, these matching strategies often pay much attention to the distance learning of the abstract features without taking the spatial stuctural and semantic correspondence learning in consideration.  

Recently, the efforts which employ deep convolutional architectures to deal with the task of person Re-ID have shown a remarkable improvement over the approaches based on the hand-craft features. For example, the patch-based methods~\cite{Alpher16}~\cite{Alpher27} perform patch-wise distance measurement to obtain the spatial relationship. Part-based methods~\cite{Alpher19} divide one person into some equal parts and jointly perform body-wise and part-wise correspondence learning based on the assumption that the pedestrian keeps upright in general. Some efforts~\cite{Alpher06} try to capture the semantic and structural correlation using deep convolution networks, which have promising results on the challenging datasets. To improve the performance of feature extraction, the triplet learning frameworks~\cite{Alpher20}~\cite{Alpher21}~\cite{Alpher22} which employ triplet training examples and the triplet loss function to learn fine grained image are also proposed.



\section{Our Architecture}
Figure \ref{fig:architecture} illustrates our network's architecture. The proposed architecture extracts color-texture and the mid-level semantic-components representation for a pair of input person images. With the features mentioned above, two pyramid matching modules are employed to learn the correspondence for the color-texture distributions and semantic-components, respectively, and to output the correspondence representations. Finally, we fuse the correspondence representations utilizing two fully connected layers and employ softmax activations to compute the final decision which indicates the probability that the image pair depicts the same person. The details of the architecture is explained in the following subsections.

\subsection{Semantic-Components (SC) Images Representation}
\begin{figure}
\centering
\includegraphics[width=0.45\textwidth]{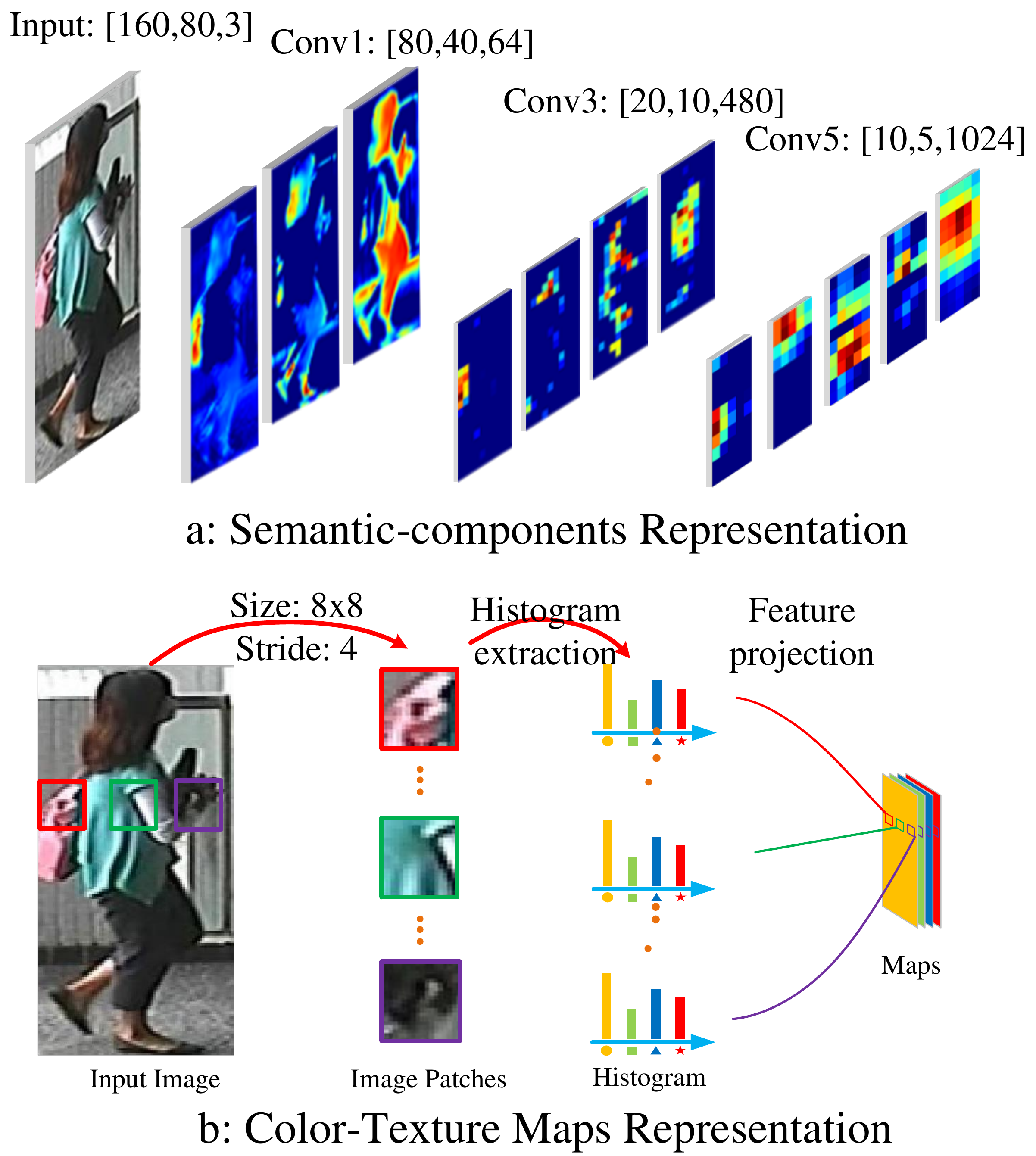}
\caption{a: Visualization of features for the ImageNet-pretrained GoogLeNet network, which is finetuned on the CUHK03 dataset. b: Illustration of the proposed  Color-Texture feature Maps (CTM) extraction. }
\label{fig:visualization}
\end{figure}

As discussed previously, there exists a set of intrinsic latent semantic components (e.g., head, front and back upper body, belongings)  in a person image, which are robust to the variations of views and background change. With these semantic representations for the images, we are able to learn the correspondence between the image pair. The well-known ImageNet-pretrained deep convolutional frameworks (like AlexNet, GoogLeNet, ResNet, etc)~\cite{Alpher09}~\cite{Alpher10}~\cite{Alpher08} have been widely used to project the RGB space to the semantic-aware space. The previous efforts~\cite{Alpher42} have also verified that the mid-level feature maps of the frameworks represent the semantic-components for one object. In our architecture, we extract these semantic-components with two parameter-shared GoogLeNets for a pair of person images. Figure \ref{fig:visualization}a shows the visualization of every block's responses in GoogLeNet finetuned on the Re-ID dataset CUHK03. We observe that the original person images are decomposed into many semantic-components (bag, head, etc.). The responses of low layers like Conv1 depict the particular components apparently and the high layers' responses like Conv5 layer look abstract but still keep the shape and spatial location. For notational simplicity, we refer to the convNet as a function $f_{CNN}( \boldsymbol X;  \boldsymbol \theta)$, that takes $ \boldsymbol X$ as input and $ \boldsymbol \theta$ as parameters. The GoogLeNets output 1024 feature maps with size $10 \times 5$ respectively as the representations of images for an input pair of images resized to $160\times80$ from two cameras, A and B. We denote this process as follows:
\begin{equation}
\label{equ:imgPre}
\{\boldsymbol R^A_{sc}, \boldsymbol R^B_{sc}\}=\{f_{CNN}(\boldsymbol I^A; \boldsymbol \theta^1_{sc}), f_{CNN}(\boldsymbol I^B; \boldsymbol \theta^1_{sc})\}
\end{equation}
where $\boldsymbol R^A_{sc}$ and $\boldsymbol R^B_{sc}$ denote the SC representation of images $\boldsymbol I^A$ and $\boldsymbol I^B$ separately. $\boldsymbol \theta^1_{sc}$ are the shared parameters. 

\subsection{Color-Texture Maps (CTM) Images Representation}



The existing methods often extract the color-texture features for images by computing the histgrams of color channels within a partitioned horizontal stripe, which works under the assumption of slight vertical misalignment, and only consider the pose variations on horizontal dimension. These methods also ignore the spatial structure information. To address these problems and represent the color spatial distributions, we propose to use sliding windows to describe local color details for a person image and construct the spatial feature maps instead of feature vectors. RGB and HSV channels are the basic color characteristics for images. The Scale Invariant Local Ternary Pattern (SILTP)~\cite{Alpher37} descriptor is an improved operator over the well-known Local Binary Pattern (LBP)~\cite{Alpher03} and an invariant texture description for illumination. Specifically, we use a subwindow size of  $8\times8$, with an overlapping step of $4$ pixels to locate local patches in the input $160\times80$ images. Within each subwindow, we extract a 24-bin RGB histogram, a 24-bin HSV histogram and a 16-bin SILTP histogram ($SILTP^{0.3}_{4,3}$). These resulting histogram-bins computed from all subwindows are then projected to the feature maps with size $40\times 20$. Figure \ref{fig:visualization}b shows the procedure of the proposed CTM extraction. 

With the extracted CTM,  we employ the parameters-shared convolution networks constructed with three convolution layers and two max-pooling layers to generate the color-texture representation with spatial size $10\times5$ consistent with the SC representation. We denote the representation above as $\boldsymbol R^A_{ctm}$ and $\boldsymbol R^B_{ctm}$ for images $\boldsymbol I^A$ and $\boldsymbol I^B$  respectively with the shared parameters $\boldsymbol \theta^1_{ctm}$:
\begin{equation}
\label{equ:CTMpre}
\{\boldsymbol R^A_{ctm}, \boldsymbol R^B_{ctm}\}=\{f_{CNN}(\boldsymbol I^A; \boldsymbol \theta^1_{ctm}), f_{CNN}(\boldsymbol I^B; \boldsymbol \theta^1_{ctm})\}
\end{equation}

\subsection{Pyramid Matching Module with Atrous Convolution}

\begin{figure}
\includegraphics[width=0.45\textwidth]{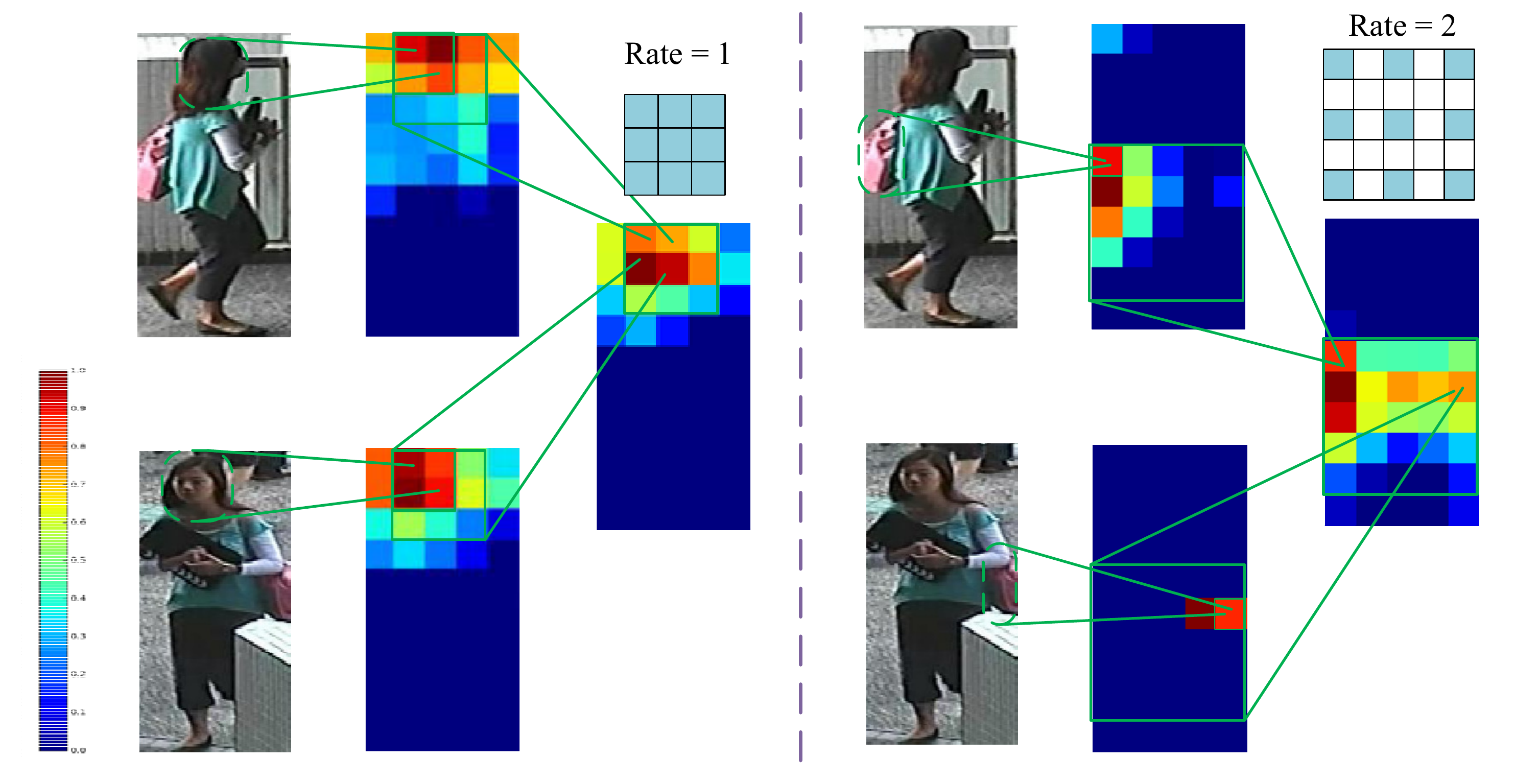}
\caption{
Illustration of correspondence learning with pyramid matching module. Left: the component ``head'' has the similiar spatial location. Right: the component ``bag'' has the completely different shape and location. We match the components above by computing their responses in one window and the convolutions with multi-scale field-of-view are robust to the misalignment and variation of scale caused by viewpoint changes.
}
\label{fig:fusion_layer}
\end{figure}
In this work, we represent the semantic-components of person images with the mid-level feature maps of GoolgLeNet, which still preserve the original shape and releative spatial location. Therefore, the variations of the spatial scale and location misalignment caused by viewpoint changes remain significant on the image representation. As shown in Figure \ref{fig:fusion_layer}, the same bag belonging to the same person is located on the right side of one image but on the left side of the other image. The previous efforts~\cite{Alpher06} address this problem by decreasing the distance for the same semantic components from two images with max-pooling layers. This strategy is effective but loses the spatial information. 

\begin{figure}
\includegraphics[width=0.47\textwidth]{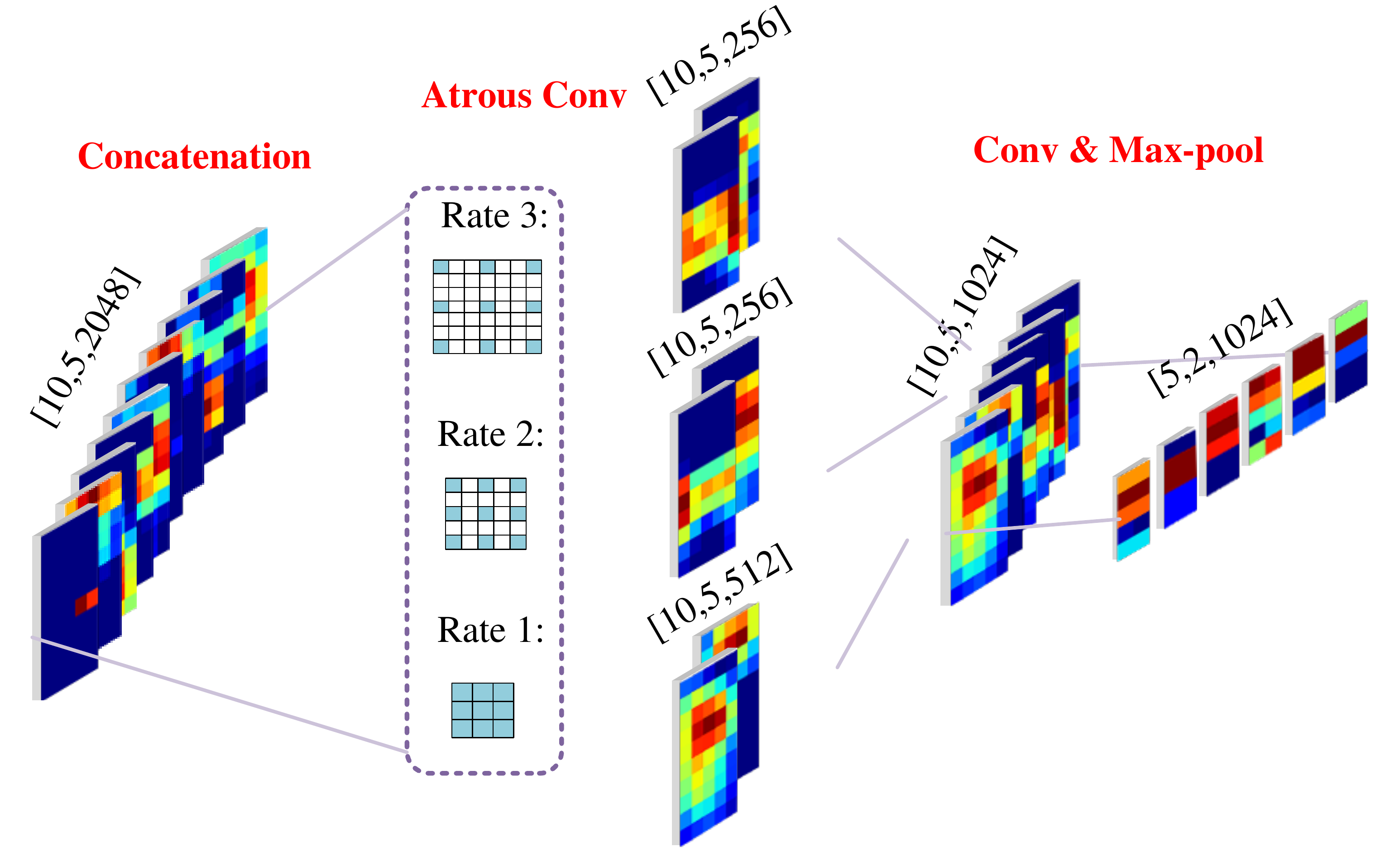}
\caption{
Illustration of the pyramid matching module.
}
\label{fig:ppm_structure}
\end{figure}

We employ atrous convolution to address this issue above. By introducing zeroes between the consecutive filter values, the atrous convolution computes the correspondences of the same semantic-components without decreasing their resolutions. Another challenge is that different semantic-components have the different scale of variations and misalignments. To address the scale invariance, we employ multi-rate atrous convolutions to construct pyramid matching module based on pyramid matching strategy to adaptively learn the correspondence for the semantic-components with multi-scale misalignments. Considering the size of feature maps, the pyramid matching module includes three branches $3\times3$ atrous convolution with rate 1, 2 and 3, which provides the field-of-view with size $3 \times 3$, $5 \times 5$, $7 \times 7$ respectively. Figure~\ref{fig:ppm_structure} shows the structure of this module and in Figure~\ref{fig:fusion_layer} two examples whose correspondences are learned with the rate1 and rate2 atrous convolutions respectively, are given to illustrate how this module works. With the images' concatenated SC representation $\{\boldsymbol R_{sc}^A,\boldsymbol R_{sc}^B\}$, the proposed module computes the correspondence distribution denoted as $\boldsymbol S_{sc}^{p} = \{ \boldsymbol S_{sc}^{r=1}$, $\boldsymbol S_{sc}^{r=2}$, $\boldsymbol S_{sc}^{r=3} \}$, in which the value of each location $(i, j)$ indicates the correspondence probability at that location. $r$ is the rate of atrous convolution. We formulate this matching strategy as follows:
\begin{align}
\label{equ:ppm}
 \boldsymbol S_{sc}^{p} ={} & \{\boldsymbol S_{sc}^{r=1}, \boldsymbol S_{sc}^{r=2}, \boldsymbol S_{sc}^{r=3}\} \notag \\
={} & \{f_{CNN}(\{\boldsymbol R_{sc}^A,\boldsymbol R_{sc}^B\}; \{\boldsymbol \theta^2_1, \boldsymbol \theta^2_2, \boldsymbol \theta^2_3\}_{sc}\} \notag \\
={} & \{f_{CNN}(\{\boldsymbol R_{sc}^A,\boldsymbol R_{sc}^B\}; \boldsymbol \theta_{sc}^2\}\} 
\end{align}
where $\boldsymbol \theta_{sc}^2 = \{ \boldsymbol \theta^2_1, \boldsymbol \theta^2_2, \boldsymbol \theta^2_3 \}_{sc}$ denotes the parameters of our module for SC representation. $\boldsymbol \theta^2_r(r=1,2,3)$ are the parameters of the matching branch with rate $r$.

We fuse the concatenated correspondence maps $\bm S_{sc}^{p}$ with learned parameters $\boldsymbol \theta_{sc}^3$, which indicate the weights of different matching branches, and output the fused correspondence representation. Inspired by \cite{Alpher06}, we further downsample the representation by max-pooling so as to preserve the most discriminative correspondence information and align it in a larger region. Finally, we obtain the correspondence representation $\boldsymbol S_{sc}^{f}$:
\begin{align}
\label{equ:weights}
\boldsymbol S_{sc}^{f}  = {}& f_{CNN}(\{\boldsymbol S_{sc}^{r=1}, \boldsymbol S_{sc}^{r=2}, \boldsymbol S_{sc}^{r=3}\}; \boldsymbol \theta_{sc}^3)  \notag \\
= {}& f_{CNN}(\{\boldsymbol R_{sc}^A,\boldsymbol R_{sc}^B\}; \boldsymbol \theta_{sc}^2, \boldsymbol \theta_{sc}^3\} 
\end{align}

Based on the same motivation and principle, we learn the correspondence of color-texture distributions of the person's attributes (e.g.clothes, hairs) with another standalone pyramid matching module. With the images' concatenated CTM representation  $\{\boldsymbol R_{ctm}^A,\boldsymbol R_{ctm}^B\}$, We obtain the correspondence representation as follows:
\begin{align}
\vspace{-1ex}
\label{equ:descriptors}
\boldsymbol S_{ctm}^{f}  = f_{CNN}(\{\boldsymbol R^A_{ctm},\boldsymbol R^B_{ctm}\}; \boldsymbol \theta^2_{ctm}, \boldsymbol \theta^3_{ctm}\} 
\end{align}
where $\boldsymbol \theta^2_{ctm}$ and $\boldsymbol \theta^3_{ctm}$ denote the parameters of pyramid matching module for CTM representation.

\subsection{The Unified Framework and Learning}
The correspondence representations $\bm S_{sc}^{f}$ and $\bm S_{ctm}^{f}$ are fused to the correspondence descriptor of size 1024 by using two fully connected layers. We pass the correspondence descriptor to another fully connected layer containing two softmax units. The probability that the two images in the pair, $\bm I^A$ and $\bm I^B$, are of the same person with softmax activations computed on the units above is denoted as:
\begin{equation}
\label{equ:softmax}
p = \frac {exp(\bm S_1(\bm S_{sc}^{f}, \bm S_{ctm}^{f}; \bm \theta^4))}{exp(\bm S_0(\bm S_{sc}^{f}, \bm S_{ctm}^{f}; \bm \theta^4))+exp(\bm S_1(\bm S_{sc}^{f}, \bm S_{ctm}^{f}; \bm \theta^4))}
\end{equation}
where $\bm S_0(\bm S_{sc}^{f}, \bm S_{ctm}^{f};\bm \theta^4)$ and $\bm S_1(\bm S_{sc}^{f}, \bm S_{ctm}^{f};\bm \theta^4)$ are the softmax units for $\bm S(\bm S_{sc}^{f}, \bm S_{ctm}^{f}\bm \theta^4)$.

We reformulate the proposed framework as a unified deep convolution framework based on Eqs.\ref{equ:imgPre} - \ref{equ:weights} :    
\begin{align}
\label{equ:fuse} 
& S(\bm S_{sc}^{f}, \bm S_{ctm}^{f}, \bm \theta^4) \notag \\
={} & f_{CNN}(\{\bm I^A,\bm I^B\}; \{\{\bm \theta^3, \{\bm \theta^2_r\},\bm \theta^1 \}_{sc}; \notag\\
{} & \{\bm \theta^3, \{\bm \theta^2_r\},\bm \theta^1 \}_{ctm};\bm \theta^4 \}) \notag \\
={} & f_{CNN}(\{\bm I^A,\bm I^B\}; \bm \theta)
\end{align}
where $\bm \theta = \{ \{\bm \theta^1, \{\bm \theta^2_r\}, \bm \theta^3\}_{sc}; \{\bm \theta^1, \{\bm \theta^2_r\}, \bm \theta^3\}_{ctm};\bm \theta^4 \}$, and $r=1,2,3$.


We minimize the widely used cross-entropy loss to optimize the network as Eq.\ref{equ:fuse} over a training set of $N$ pairs using stochastic gradient descent. $l_n$ is the 1/0 label for the input pair depicting whether the same person or not. With this unified network, the processes of discriminative image representation learning and cross-person correspondence learning are optimized jointly to make the image representation optimal to this task. 
\begin{align}
\label{equ:loss} 
\bm L = - \frac{1}{N} \sum^N_{n=1} [ l_n \log p_n + (1-l_n) \log (1-p_n) ]
\end{align}

By setting $\{\bm \theta^1, \{\bm \theta^2_r\}, \bm \theta^3\}_{sc} = \bm 0$ or $\{\bm \theta^1, \{\bm \theta^2_r\}, \bm \theta^3\}_{ctm} = \bm 0$, we construct two independent convNets named SC-PPMN and CTM-PPMN which focus on semantic-components correspondence learning and color-texture distributions correspondence learning  respectively. These two convNets are denoted as Eq.\ref{equ:unified_sc} and Eq.\ref{equ:unified_ctm} optimized with $\bm L_{sc}$ and $\bm L_{ctm}$ represented in Eq.\ref{equ:loss} respectively.

\begin{align}
\label{equ:unified_sc} 
& S_{sc}(\bm S_{sc}^{f}, \bm \theta_{sc}^4) \notag \\
={} & f_{CNN}(\{\bm I^A,\bm I^B\}; \{\{\bm \theta^3, \{\bm \theta^2_r\},\bm \theta^1\}_{sc};\bm \theta_{sc}^4\}) \notag \\
={} & f_{CNN}(\{\bm I^A,\bm I^B\}; \bm \theta_{sc})
\end{align}
\begin{align}
\label{equ:unified_ctm} 
& S_{ctm}(\bm S_{ctm}^{f}, \bm \theta_{ctm}^4) \notag \\
={} & f_{CNN}(\{\bm I^A,\bm I^B\}; \{\{\bm \theta^3, \{\bm \theta^2_r\},\bm \theta^1 \}_{ctm};\bm \theta_{ctm}^4\}) \notag \\
={} & f_{CNN}(\{\bm I^A,\bm I^B\}; \bm \theta_{ctm})
\end{align}

\section{Experiments}
\subsection{Datasets and Protocol}

We evaluate the proposed architecture and compare our results with those of the state-of-the-art approaches on six person Re-ID datasets, namely CUHK03~\cite{Alpher27}, CUHK01~\cite{Alpher28}, VIPeR~\cite{Alpher29}, PRID450s~\cite{Alpher38}, i-LIDS~\cite{Alpher39} and PRID2011~\cite{Alpher40}. All the approaches are evaluated with Cumulative Matching Characteristics (CMC) curves by single-shot results, which characterize a ranking result for every image in the gallery given the probe image. Our experiments are conducted on the datasets with 10 random training and the average results are presented.  We conduct the experiments on SC-PPMN, CTM-PPMN and MC-PPMN to learn the correspondence for two person images with the CTM features, SC features and the fused features, respectively. We report the experimental results and analyze the performances of CTM features and SC features.

\begin{table}[!htbp]
\centering
\label{table:dataset}
\caption{Datasets and settings in our experiments.}
\vspace{1ex}
\scalebox{0.7}{

\begin{tabular}{|c|c|c|c|c|c|c|}
\hline
 Dataset & CUHK03 & CUHK01 & VIPeR & PRID450s & i-LIDS & PRID2011\\
\hline
 identities & 1360 & 971 & 632 & 450 & 119 & 385/749\\
 images & 13164 & 3884 & 1264 & 900 & 479 & 1134\\
 views & 2 & 2 & 2 & 2 & 2 & 2\\
 train IDs & 1160 & 871;485 & 316 & 225 & 59 & 100\\
 test IDs & 100 & 100;486 & 316 & 225 & 59 & 100\\
\hline
\end{tabular}} 
\end{table}

Table \ref{table:dataset} lists the description of each dataset and our experimental settings with the training and testing splits. The CUHK03 dataset provides two settings named labelled setting with the manually annotated pedestrian bounding boxes and detected settings with automatically generated bounding boxes in which possible misalignments and body part missing are introduced for a more realistic setting. In this paper, the evaluation results on both labelled and detected settings are reported. For the CUHK01 dataset, we report results on two different settings: 100 test IDs, and 486 test IDs. The VIPeR and PRID450s dataset are relatively small datasets and only contain one image per person in each view. i-LIDS dataset is constructed from video images shooting a busy airport arrival hall and contains 479 images from 119 persons, in which each person has four images in average. PRID2011 dataset consists of images captured by two static surveillance cameras, in which views A and B contain 385 and 749 persons, respectively, with 200 persons appearing in both views. Following the procedure described in ~\cite{Alpher33} for evaluation on the test set, view A is used for the probe set (100 person IDs) and view B is used for the gallery set, which contains all images of the view B (649 person IDs) except the 100 training samples.

\subsection{Training the Network}
The proposed architecture is implemented on the widely used deep learning framework Caffe~\cite{Alpher30} with an NVIDIA TITAN X GPU. We use stochastic gradient descent(SGD) for updating the weights of the network. The parameters for training SC-PPMN, CTM-PPMN and MC-PPMN are listed in Table \ref{table:para}. We start with a base learning rate and gradually decrease it as the training progresses using a polynomial decay policy: $ \eta^{i} = \eta^{0}(1-\frac{i}{max\_iter})^p$, where $p = 0.5$, $i$ is the current mini-batch iteration and $max\_iter$ is the maximum iteration. We train the MC-PPMN model by fixing the parameters of the pre-trained SC-PPMN and CTM-PPMN models.  

\textbf{Data Augmentation}. To make the model robust to the image translation variation and to enlarge the data set, we sample 5 images around the image center, with translation drawn
from a uniform distribution in the range $[-8,8]\times[-4,4]$ for an original image of size $160\times80$.

\textbf{Hard Negative Mining (hnm)}. In fact, the negative pairs are far more than the positive pairs, which can lead to data imbalance. Also, in these negative pairs, there still exist some scenarios that are hard to distinguish. To address these difficuties, we sample the hard negative piars for retraining our network following the way in ~\cite{Alpher16}.

\begin{table}[!htbp]
\centering
\caption{The parameters for training.}
\label{table:para}
\vspace{1ex}
\scalebox{0.8}{
\begin{tabular}{|c|c|c|c|}
\hline
 Parameters & SC-PPMN & CTM-PPMN & MC-PPMN \\
\hline
 Training Time (hours) & 40-48 & 16 & 10 \\
 Maximum Iteration & 160K & 30K & 10K \\
 Batch Size & 100 & 800 & 150 \\
 Momentum & 0.9 & 0.9 & 0.9 \\
 Weight Decay & 0.0002 & 0.0002 & 0.0002 \\
 Base Learning Rate & 0.01 & 0.1 & 0.0001 \\
\hline
\end{tabular}} 
\end{table}

\begin{table}
\centering
\caption{Comparison of state-of-the-art results on labelled and detected CUHK03 dataset with 100 test IDs. The cumulative matching scores (\%) at rank 1, 5, and 10 are listed.}
\label{table:CUHK03}
\vspace{1ex}
\scalebox{0.75}{
\begin{tabular}{c|ccc|ccc}
\hline
\multirow{2}*{Methods} & 
\multicolumn{3}{c|}{labelled CUHK03} &
\multicolumn{3}{c}{detected CUHK03}  \\ 
\cline{2-7}
 & r=1 & r=5 & r=10 & r=1 & r=5 & r=10  \\ \hline
 KISSME & 14.17 & 37.46 & 52.20 & 11.70 & 33.45 & 45.69  \\
 LMNN  & 7.29 & 19.64 & 30.74 & 6.25 & 17.87 & 26.60 \\
 LSSCDL & 57.00 & - & - & 51.20 & - & -  \\
 LOMO+LSTM & - & - & - & 57.30 & 80.10 & 88.30 \\
 LOMO+XQDA & 52.20 & 82.23 & 92.14 & 46.25 & 78.90 & 88.55 \\ \hline
 CTM-PPMN (no hnm) & 73.52 & 95.12 & 98.56 & 68.44 & 91.50 & 96.98 \\
 CTM-PPMN (hnm) & 76.58 & 95.64 & 98.24 & 70.68 & 92.58 & 97.18 \\  \hline
 \hline
 FPNN & 20.65 & 50.94 & 67.01 & 19.89 & 49.41 & -  \\
 ImprovedDL & 54.74 & 86.50 & 93.88 & 44.96 & 76.01 & 81.85  \\
 PIE(R)+Kissme & - & - & - & 67.10 & 92.20 & 96.60  \\
 SICIR & - & - & - & 52.17 & - & - \\
 DCSL (no hnm) & 78.60 & 97.76 & 99.30 & - & - & - \\
 DCSL (hnm) & 80.20 & 97.73 & 99.17 & - & - & -  \\ 
 MTDnet & 74.68 & 95.99 & 97.47 & - & - & - \\
 JLML & 83.20 & 98.00 & 99.40 & 80.60 & \textbf{96.90} & \textbf{98.70} \\ \hline
 SC-PPMN (no hnm) & 83.20 & 97.50 & 99.25 & 77.60 & 96.10 & 98.60  \\
 SC-PPMN (hnm) & 85.50 & 98.20 & 99.50 & 80.63 & 95.62 & 98.07 \\ \hline
 \hline
 MC-PPMN (no hnm) & 84.36 & \textbf{98.56} & \textbf{99.80} & 81.28 & 96.14 & 98.54 \\
 MC-PPMN (hnm) & \textbf{86.36} & 98.54 & 99.66  & \textbf{81.88} & 96.56 & 98.58\\  \hline
\end{tabular}}
\end{table}

\subsection{Experiments Results}

We campare our proposed MC-PPMN with several methods in recent years, including both hand-craft feature based methods: ITML~\cite{Alpher42}, LMNN~\cite{Alpher13}, KISSME~\cite{Alpher14}, LOMO+XQDA~\cite{Alpher32}, LSSCDL~\cite{Alpher31}, LOMO+LSTM~\cite{Alpher19}; and DCNN feature based methods: FPNN~\cite{Alpher27}, ImprovedDL~\cite{Alpher16}, Single-Image and Cross-Images Representation learning (SICIR)~\cite{Alpher22}, TCP~\cite{Alpher33}, DCSL~\cite{Alpher06}, Pose Invariant Embedding (PIE(R)+Kissme)~\cite{Alpher34}, MTDnet (including MTDnet-cross)~\cite{Alpher43}, JLML~\cite{Alpher41}. We report the evaluation results as shown in Table~\ref{table:CUHK03} - Table~\ref{table:i-LIDS}.

\textbf{Comparisons on CUHK03 dataset}.
We conduct the experiments on both labelled and detected CUHK03 datasets. From Table \ref{table:CUHK03}, we see that our proposed approach achieves the better results than the state-of-the-art methods. On the labelled dataset, our method outperforms the next best method by an improvement of 3.16\% (86.36\% vs. 83.20\%). On the detected dataset, the performance is reduced by the misalignment and incompleteness caused by the detector. However, the proposed method still achieves an improvement 1.28\% over the next best method (81.88\% vs. 80.60\%).
\begin{table}[!htbp]
\centering
\caption{Comparison of state-of-the-art results on CUHK01 dataset with 100 test IDs and 486 test IDs. The cumulative matching scores (\%) at rank 1, 5, and 10 are listed.}
\label{table:CUHK01}
\vspace{1ex}
\scalebox{0.75}{
\begin{tabular}{c|ccc|ccc}
\hline
\multirow{2}*{Methods} & 
\multicolumn{3}{c|}{CUHK01(100 test IDs)} &
\multicolumn{3}{c}{CUHK01(486 test IDs)} \\ 
\cline{2-7}
 & r=1 & r=5 & r=10 & r=1 & r=5 & r=10 \\ \hline
 KISSME  & 29.40 & 60.18 & 74.44 & - & - & - \\
 LMNN  & 21.17 & 48.51 & 62.98 & 13.45 & 31.33 & 42.25\\
 LSSCDL & 65.97 & 48.51 & 62.98 & - & - & - \\ \hline
 CTM-PPMN (no hnm) & 71.18 & 91.94 & 96.54 & 48.01 & 75.91 & 84.34 \\
 CTM-PPMN (hnm) & 73.74 & 92.32 & 98.18 & 53.57 & 79.32 & 87.13 \\  \hline
 \hline
 FPNN & 27.87 & 59.64 & 73.53 & - & - & - \\
 ImprovedDL  & 65.00 & 89.00 & 94.00 & 47.53 & 71.60 & 80.25 \\
 SICIR  & 71.80 & - & - & - & - & - \\
 TCP & - & - & - & 53.70 & 84.30 & 91.00 \\
 MTDnet-cross & 78.50 & 96.50 & 97.50 & - & - & - \\
 DCSL (no hnm) & 88.00 & 96.90 & 98.10 & - & - & - \\
 DCSL (hnm) & 89.60 & 97.80 & 98.90 & 76.54 & 94.24 & 97.49 \\ \hline
 SC-PPMN (no hnm) & 92.10 & 99.50 & 99.95 & - & - & -  \\
 SC-PPMN (hnm) & 93.10 & 98.80 & 99.80 & 77.16 & 92.80 & 97.53  \\ \hline
 \hline
 MC-PPMN (no hnm)  & 92.32 & 98.68 & 99.60 & - & - & - \\
 MC-PPMN (hnm)   & \textbf{93.45} & \textbf{99.62} & \textbf{99.98} & \textbf{78.95} & \textbf{94.67} & \textbf{97.64} \\ \hline
\end{tabular}}
\end{table}

\textbf{Comparisons on CUHK01 dataset}. 
Table \ref{table:CUHK01} illustrates the top recognition rate on CUHK01 dataset with 100 test IDs and 486 test IDs. We see that our proposed method achieves the best recognition rate of 93.45\% (rank-1), 99.62\% (rank-5) and 99.98\% (rank-10) (vs. 89.60\%, 96.90\% and 99.98\% respectively by the next best method) with 100 test IDs. For the setting with 486 test IDs, only 485 identities and half positive samples are left for training which make it challenging for our proposed deep architecture to converge. Following the way in \cite{Alpher06}, we finetune the network for CUHK01 with the pre-trained model on CUHK03 and achieve an improvement of 2.41\%(78.95\% vs. 76.54\%) on rank-1 recognition rate. 

\begin{table}
\centering
\caption{Comparison of state-of-the-art results on VIPeR and PRID450S datasets.The cumulative matching scores (\%) at rank 1, 5, and 10 are listed.}
\label{table:VIPeR}
\vspace{1ex}
\scalebox{0.70}{
\begin{tabular}{c|ccc|ccc}
\hline
\multirow{2}*{Methods} & 
\multicolumn{3}{c|}{VIPeR} &
\multicolumn{3}{c}{PRID450s} \\
\cline{2-7}
 & r=1 & r=5 & r=10 & r=1 & r=5 & r=10 \\ \hline
 KISSME & 19.60 & 48.00 & 62.20 & 15.0 & - & 39.0  \\
 LSSCDL & 42.66 & - & 84.27 &  60.49 & - & 88.58 \\
 LOMO+LSTM  & 42.40 & 68.70 & 79.40 & - & - & -\\
 LOMO+XQDA & 40.00 & 68.13 & 80.51 & 61.42 & - & 90.84 \\ \hline
 CTM-PPMN & 32.12 & 64.24 & 80.38 & 28.98 & 59.47 & 73.60 \\ \hline
 \hline
 ImprovedDL & 34.81 & 63.61 & 75.63 & 34.81 & 63.72 & 76.24 \\
 PIE(R) & 27.44 & 43.01 & 50.82 & - & - & -\\
 SICIR & 35.76 & - & - & - & - & -\\
 TCP & 47.80 & 74.70 & 84.80 & - & - & -\\
 DCSL & 44.62 & 73.42 & 82.59 & - & - & - \\ 
 JLML & \textbf{50.20} & 74.20 & 84.30  & - & - & -\\ \hline
 SC-PPMN & 45.82 & 74.68 & 86.08 & 52.08 & 82.58 & 88.36 \\ \hline
 \hline
 MC-PPMN & 50.13 & \textbf{81.17} & \textbf{91.46} & \textbf{62.22} & \textbf{84.00} & \textbf{93.56}\\ \hline
\end{tabular}}
\end{table}

\textbf{Comparisons on VIPeR and PRID450s dataset}.
Following \cite{Alpher16}, we pre-train the network using CUHK03 and CUHK01 datasets, and fine-tune on the training set of VIPeR and PRID450s. As shown in the Table \ref{table:VIPeR}, the proposed MC-PPMN is better than the state-of-the-art method in all the cases except the rank-1 recognition rate for VIPeR dataset, while is comparable with the best competing method JLML.

\begin{table}
\centering
\caption{Comparison of state-of-the-art results on i-LIDS and PRID2011 datasets. The cumulative matching scores (\%) at rank 1, 5, and 10 are listed.}
\label{table:i-LIDS}
\vspace{1ex}
\scalebox{0.70}{
\begin{tabular}{c|ccc|ccc}
\hline
\multirow{2}*{Methods} & 
\multicolumn{3}{c|}{i-LIDS} &
\multicolumn{3}{c}{PRID2011} \\
\cline{2-7}
 & r=1 & r=5 & r=10 & r=1 & r=5 & r=10\\ \hline
 ITML & 29.00 & 54.00 & 70.50 & 12.00 & - & 36.00 \\
 KISSME & - & - & - & 15.00 & - & 39.00  \\
 LMNN & 28.00 & 53.80 & 66.10 & 10.00 & - & 30.00 \\ \hline
 CTM-PPMN & 44.17 & 73.31 & 85.02  & 12.00 & 32.00 & 42.00 \\  \hline
 \hline
 TCP  & 60.40 & 82.70 & 90.70 & 22.00 & 47.00 & 57.00 \\ 
 MTDnet & 57.8 & 78.61 & 87.28 & 32.00 & 51.00 & 62.00 \\
 \hline
 SC-PPMN & 54.80 & 81.92 & 92.32 & 32.00 & 53.00 & 63.00 \\ \hline
 \hline
 MC-PPMN & \textbf{62.69} & \textbf{84.80} & \textbf{93.31} & \textbf{34.00} & \textbf{60.00} & \textbf{69.00}\\ \hline
\end{tabular}}
\end{table}

\textbf{Comparisons on i-LIDS and PRID2011 datasets}.
We also conduct experiments on the i-LIDS dataset and PRID2011 dataset. Table \ref{table:i-LIDS} shows our results. For both datasets, MC-PPMN achieves the best rank-1, rank-5 and rank-10 recognition rates, which demonstrate the effectiveness of the proposed method for the small training set. 

\begin{table}
\centering
\caption{The improvement of the fused correspondence representations for rank-1 recognition rates on the experimental datasets.}
\label{table:enhancement}
\vspace{1ex}
\scalebox{0.68}{
\begin{tabular}{c|ccc|c}
\hline
 Dataset & CTM-PPMN & SC-PPMN & MC-PPMN & Improvement \\ \hline
 CUHK03(labelled) & 76.58 & 85.50 & 86.36 & 0.86 \\
 CUHK03(detected) & 70.68 & 80.63 & 81.88 & 1.25 \\
 CUHK01(100 test IDs) & 73.74 & 93.10 & 93.45 & 0.35 \\
 CUHK01(486 test IDs) & 53.57 & 77.16 & 78.95 & 1.79 \\
 VIPeR & 32.12 & 45.82 & 50.13 & 4.31 \\
 PRID450s & 28.98 & 52.08 & 62.22 & 10.14\\
 i-LIDS & 44.17 & 54.80 & 62.69 & 7.89\\
 PRID2011 & 12.00 & 32.00 & 34.00 & 2.00 \\ \hline
\end{tabular}}
\end{table}

\textbf{The effect of fusion for the correspondence representations}.  Camparing with the experimental results by learning the correspondence for two person images with CTM features and SC features, respectively, Table \ref{table:enhancement} shows the improvement on the rank-1 recognition rates with the fusion for the correspondence representations. For CUHK03 and CUHK01 datasets, we achieve the absolute gain about 1.00\% and for the other datasets, we can see the absolute gain over 2.00\%. Especially, the proposed method achieve 10.14\% improvement on the rank-1 recognition rate. The results above demonstrate the effectiveness of fusion for the correspondence representations, which is obvious on the small datasets.      

\textbf{The effect of hard negative mining}. We also report the results of both our model with hnm and without hnm as shown in Table \ref{table:CUHK03} and \ref{table:CUHK01}. We can see the absolute gain about 1.00\% compared with the same model without hnm.

\section{Conclusion}
In this paper, we have developed a novel multi-channel deep convolutional architecture for person re-identification. We employ deep convNets to map person's semantic components and color-texture distributions to the required feature space. Based on the learned deep features and a pyramid matching strategy, we learn their correspondence representations and fuse them together to perform the re-identification task. The effectiveness and promise of our method is demonstrated by extensive evaluations on various datasets. The results have shown that our method has a remarkable improvement over the competing models.

\section{ Acknowledgments}
\noindent This work was supported by National Key R\&D Program of China (No. 2017YFB1002400), National Natural Science Foundation of China (No. 61702448, 61672456), the Key R\&D Program of Zhejiang Province (No. 2018C03042), the Fundamental Research Funds for the Central Universities (No. 2017QNA5008, 2017FZA5007). X. Li was also supported in part by the National Natural Science Foundation of China under Grant U1509206 and Grant 61472353, and the Alibaba-Zhejiang University Joint Institute of Frontier Technologies.


\bibliography{egbib}

\begin{thebibliography}{}

\bibitem[\protect\citeauthoryear{Ahmed, Jones, and Marks}{2015}]{Alpher16}
Ahmed, E.; Jones, M.; and Marks, T.~K.
\newblock 2015.
\newblock An improved deep learning architecture for person re-identification.
\newblock In {\em CVPR},  3908--3916.

\bibitem[\protect\citeauthoryear{Chen \bgroup et al\mbox.\egroup
  }{2016}]{Alpher23}
Chen, L.~C.; Papandreou, G.; Kokkinos, I.; Murphy, K.; and Yuille, A.~L.
\newblock 2016.
\newblock Deeplab: Semantic image segmentation with deep convolutional nets,
  atrous convolution, and fully connected crfs.
\newblock {\em IEEE Transactions on Pattern Analysis and Machine Intelligence}
  PP(99):1--1.

\bibitem[\protect\citeauthoryear{Chen \bgroup et al\mbox.\egroup
  }{2017}]{Alpher43}
Chen, W.; Chen, X.; Zhang, J.; and Huang, K.
\newblock 2017.
\newblock A multi-task deep network for person re-identification.
\newblock In {\em AAAI},  3988--3994.

\bibitem[\protect\citeauthoryear{Cheng \bgroup et al\mbox.\egroup
  }{2016a}]{Alpher20}
Cheng, D.; Gong, Y.; Zhou, S.; Wang, J.; and Zheng, N.
\newblock 2016a.
\newblock Person re-identification by multi-channel parts-based cnn with
  improved triplet loss function.
\newblock In {\em CVPR},  1335--1344.

\bibitem[\protect\citeauthoryear{Cheng \bgroup et al\mbox.\egroup
  }{2016b}]{Alpher33}
Cheng, D.; Gong, Y.; Zhou, S.; Wang, J.; and Zheng, N.
\newblock 2016b.
\newblock Person re-identification by multi-channel parts-based cnn with
  improved triplet loss function.
\newblock In {\em CVPR},  1335--1344.

\bibitem[\protect\citeauthoryear{Davis \bgroup et al\mbox.\egroup
  }{2007}]{Alpher42}
Davis, J.~V.; Kulis, B.; Jain, P.; Sra, S.; and Dhillon, I.~S.
\newblock 2007.
\newblock Information-theoretic metric learning.
\newblock In {\em ICML},  209--216.

\bibitem[\protect\citeauthoryear{Ding \bgroup et al\mbox.\egroup
  }{2015}]{Alpher21}
Ding, S.; Lin, L.; Wang, G.; and Chao, H.
\newblock 2015.
\newblock Deep feature learning with relative distance comparison for person
  re-identification.
\newblock {\em Pattern Recognition} 48(10):2993--3003.

\bibitem[\protect\citeauthoryear{Farenzena \bgroup et al\mbox.\egroup
  }{2010}]{Alpher01}
Farenzena, M.; Bazzani, L.; Perina, A.; Murino, V.; and Cristani, M.
\newblock 2010.
\newblock Person re-identification by symmetry-driven accumulation of local
  features.
\newblock In {\em CVPR},  2360--2367.

\bibitem[\protect\citeauthoryear{Gray and Tao}{2008}]{Alpher29}
Gray, D., and Tao, H.
\newblock 2008.
\newblock Viewpoint invariant pedestrian recognition with an ensemble of
  localized features.
\newblock {\em Computer Vision--ECCV}  262--275.

\bibitem[\protect\citeauthoryear{He \bgroup et al\mbox.\egroup
  }{2016}]{Alpher10}
He, K.; Zhang, X.; Ren, S.; and Sun, J.
\newblock 2016.
\newblock Deep residual learning for image recognition.
\newblock In {\em CVPR},  770--778.

\bibitem[\protect\citeauthoryear{Hirzer \bgroup et al\mbox.\egroup
  }{2011}]{Alpher40}
Hirzer, M.; Beleznai, C.; Roth, P.~M.; and Bischof, H.
\newblock 2011.
\newblock Person re-identification by descriptive and discriminative
  classification.
\newblock In {\em Scandinavian Conference on Image Analysis},  91--102.

\bibitem[\protect\citeauthoryear{Jia \bgroup et al\mbox.\egroup
  }{2014}]{Alpher30}
Jia, Y.; Shelhamer, E.; Donahue, J.; Karayev, S.; Long, J.; Girshick, R.;
  Guadarrama, S.; and Darrell, T.
\newblock 2014.
\newblock Caffe: Convolutional architecture for fast feature embedding.
\newblock In {\em ACM MM},  675--678.
\newblock ACM.

\bibitem[\protect\citeauthoryear{Koestinger \bgroup et al\mbox.\egroup
  }{2012}]{Alpher14}
Koestinger, M.; Hirzer, M.; Wohlhart, P.; Roth, P.~M.; and Bischof, H.
\newblock 2012.
\newblock Large scale metric learning from equivalence constraints.
\newblock In {\em CVPR},  2288--2295.
\newblock IEEE.

\bibitem[\protect\citeauthoryear{Krizhevsky, Sutskever, and
  Hinton}{2012}]{Alpher08}
Krizhevsky, A.; Sutskever, I.; and Hinton, G.~E.
\newblock 2012.
\newblock Imagenet classification with deep convolutional neural networks.
\newblock In {\em Advances in neural information processing systems},
  1097--1105.

\bibitem[\protect\citeauthoryear{Li and Wang}{2013}]{Alpher03}
Li, W., and Wang, X.
\newblock 2013.
\newblock Locally aligned feature transforms across views.
\newblock In {\em CVPR},  3594--3601.

\bibitem[\protect\citeauthoryear{Li \bgroup et al\mbox.\egroup
  }{2014}]{Alpher27}
Li, W.; Zhao, R.; Xiao, T.; and Wang, X.
\newblock 2014.
\newblock Deepreid: Deep filter pairing neural network for person
  re-identification.
\newblock In {\em CVPR},  152--159.

\bibitem[\protect\citeauthoryear{Li, Zhao, and Wang}{2012}]{Alpher28}
Li, W.; Zhao, R.; and Wang, X.
\newblock 2012.
\newblock Human reidentification with transferred metric learning.
\newblock In {\em ACCV},  31--44.
\newblock Springer.

\bibitem[\protect\citeauthoryear{Li, Zhu, and Gong}{2017}]{Alpher41}
Li, W.; Zhu, X.; and Gong, S.
\newblock 2017.
\newblock Person re-identification by deep joint learning of multi-loss
  classification.
\newblock In {\em IJCAI},  2194--2200.

\bibitem[\protect\citeauthoryear{Liao \bgroup et al\mbox.\egroup
  }{2010}]{Alpher37}
Liao, S.; Zhao, G.; Kellokumpu, V.; Pietikainen, M.; and Li, S.~Z.
\newblock 2010.
\newblock Modeling pixel process with scale invariant local patterns for
  background subtraction in complex scenes.
\newblock In {\em CVPR},  1301--1306.

\bibitem[\protect\citeauthoryear{Liao \bgroup et al\mbox.\egroup
  }{2015}]{Alpher32}
Liao, S.; Hu, Y.; Zhu, X.; and Li, S.~Z.
\newblock 2015.
\newblock Person re-identification by local maximal occurrence representation
  and metric learning.
\newblock In {\em CVPR},  2197--2206.

\bibitem[\protect\citeauthoryear{Mignon and Jurie}{2012}]{Alpher11}
Mignon, A., and Jurie, F.
\newblock 2012.
\newblock Pcca: A new approach for distance learning from sparse pairwise
  constraints.
\newblock In {\em CVPR},  2666--2672.

\bibitem[\protect\citeauthoryear{Office}{2008}]{Alpher39}
Office, U.~H.
\newblock 2008.
\newblock i-lids multiple camera tracking scenario definition.

\bibitem[\protect\citeauthoryear{Pedagadi \bgroup et al\mbox.\egroup
  }{2013}]{Alpher12}
Pedagadi, S.; Orwell, J.; Velastin, S.; and Boghossian, B.
\newblock 2013.
\newblock Local fisher discriminant analysis for pedestrian re-identification.
\newblock In {\em CVPR},  3318--3325.

\bibitem[\protect\citeauthoryear{Roth \bgroup et al\mbox.\egroup
  }{2014}]{Alpher38}
Roth, P.~M.; Hirzer, M.; Koestinger, M.; Beleznai, C.; and Bischof, H.
\newblock 2014.
\newblock Mahalanobis distance learning for person re-identification.
\newblock In {\em Person Re-Identification}. Springer.
\newblock  247--267.

\bibitem[\protect\citeauthoryear{Szegedy \bgroup et al\mbox.\egroup
  }{2015}]{Alpher09}
Szegedy, C.; Liu, W.; Jia, Y.; Sermanet, P.; Reed, S.; Anguelov, D.; Erhan, D.;
  Vanhoucke, V.; and Rabinovich, A.
\newblock 2015.
\newblock Going deeper with convolutions.
\newblock In {\em CVPR},  1--9.

\bibitem[\protect\citeauthoryear{Varior \bgroup et al\mbox.\egroup
  }{2016}]{Alpher19}
Varior, R.~R.; Shuai, B.; Lu, J.; Xu, D.; and Wang, G.
\newblock 2016.
\newblock A siamese long short-term memory architecture for human
  re-identification.
\newblock In {\em ECCV},  135--153.
\newblock Springer.

\bibitem[\protect\citeauthoryear{Wang \bgroup et al\mbox.\egroup
  }{2016}]{Alpher22}
Wang, F.; Zuo, W.; Lin, L.; Zhang, D.; and Zhang, L.
\newblock 2016.
\newblock Joint learning of single-image and cross-image representations for
  person re-identification.
\newblock In {\em CVPR},  1288--1296.

\bibitem[\protect\citeauthoryear{Weinberger and Saul}{2009}]{Alpher13}
Weinberger, K.~Q., and Saul, L.~K.
\newblock 2009.
\newblock Distance metric learning for large margin nearest neighbor
  classification.
\newblock volume~10,  207--244.

\bibitem[\protect\citeauthoryear{Zhang \bgroup et al\mbox.\egroup
  }{2016a}]{Alpher06}
Zhang, Y.; Li, X.; Zhao, L.; and Zhang, Z.
\newblock 2016a.
\newblock Semantics-aware deep correspondence structure learning for robust
  person re-identification.
\newblock In {\em IJCAI},  3545--3551.

\bibitem[\protect\citeauthoryear{Zhang \bgroup et al\mbox.\egroup
  }{2016b}]{Alpher31}
Zhang, Y.; Li, B.; Lu, H.; Irie, A.; and Ruan, X.
\newblock 2016b.
\newblock Sample-specific svm learning for person re-identification.
\newblock In {\em CVPR},  1278--1287.

\bibitem[\protect\citeauthoryear{Zhao, Ouyang, and Wang}{2013}]{Alpher02}
Zhao, R.; Ouyang, W.; and Wang, X.
\newblock 2013.
\newblock Unsupervised salience learning for person reidentification.
\newblock In {\em CVPR},  3586--3593.

\bibitem[\protect\citeauthoryear{Zheng \bgroup et al\mbox.\egroup
  }{2017}]{Alpher34}
Zheng, L.; Huang, Y.; Lu, H.; and Yang, Y.
\newblock 2017.
\newblock Pose invariant embedding for deep person re-identification.
\newblock {\em arXiv preprint arXiv:1701.07732}.

\end{thebibliography}
\bibliographystyle{aaai}
\end{document}